\documentclass[final]{cvpr}

\usepackage{times}
\usepackage{epsfig}
\usepackage{graphicx}
\usepackage{amsmath}
\usepackage{amssymb}

\usepackage{color}
\usepackage{amsfonts}
\usepackage{balance}
\usepackage{algorithm, algorithmic}
\usepackage{breqn}
\usepackage{multirow, hhline}
\usepackage{xspace}
\usepackage{comment}
\usepackage{stfloats}

\usepackage{array}
\usepackage{ftnright}
\usepackage{comment}
\usepackage{subcaption}

\newcolumntype{C}{ >{\centering\arraybackslash} p{1.2cm} }
\newcolumntype{H}{ >{\centering\arraybackslash} p{2.8cm} }

\usepackage[pagebackref=true,breaklinks=true,colorlinks,bookmarks=false]{hyperref}

\begin{document}

\def \OURS {\textit{SpoC}}
\title{SpoC: Spoofing Camera Fingerprints}

\author{
	Davide Cozzolino\textsuperscript{1} \ \ \,\,\,
	Justus Thies\textsuperscript{2} \ \ \ \ \ \ \,\,\,
	Andreas R\"ossler\textsuperscript{2} \ \ \ \,\,\,
	Matthias Nie{\ss}ner\textsuperscript{2} \ \ \ \,\,\,
	Luisa Verdoliva\textsuperscript{1} \\
	\small \textsuperscript{1}University Federico II of Naples \ \ \ \ \ \textsuperscript{2}Technical University of Munich \ \ \ \ \ 
}

\twocolumn[{%
	\renewcommand\twocolumn[1][]{#1}%
	\maketitle
	\begin{center}
		\includegraphics[width=\linewidth]{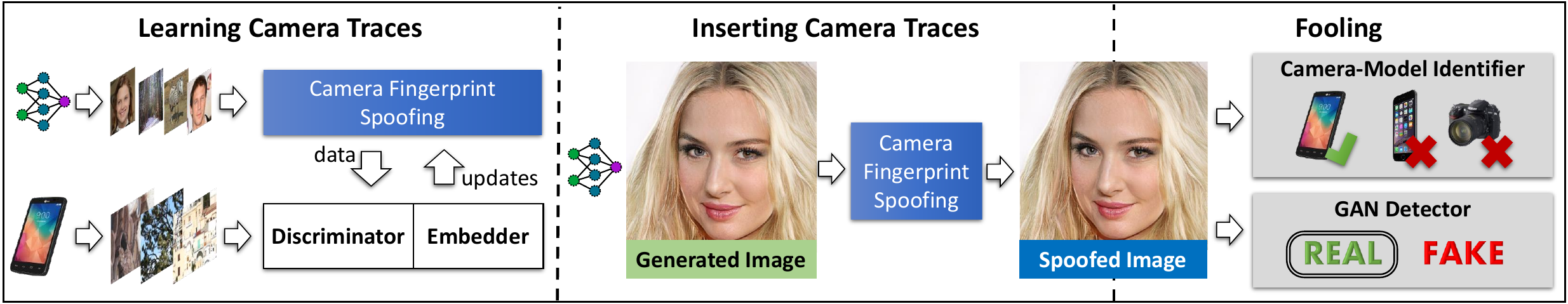}
		\captionof{figure}{
		\OURS\xspace learns to spoof camera fingerprints. It can be used to insert camera traces to a generated image. Experiments show that we can fool both camera-model identifiers and GAN detectors which were not seen during training.	}
		\label{fig:teaser}
	\end{center}
}]

\begin{abstract}

Thanks to the fast progress in synthetic media generation, creating realistic false images has become very easy. Such images can be used to wrap  “rich” fake news with enhanced credibility, spawning a new wave of high-impact, high-risk misinformation campaigns. Therefore, there is a fast-growing interest in reliable detectors of manipulated media. The most powerful detectors, to date, rely on the subtle traces left by any device on all images acquired by it. In particular, due to proprietary in-camera processes, like demosaicing or compression, each camera model leaves trademark traces that can be exploited for forensic analyses.  The absence or distortion of such traces in the target image is a strong hint of manipulation. In this paper, we challenge such detectors to gain better insight into their vulnerabilities. This is an important study in order to build better forgery detectors able to face malicious attacks. Our proposal consists of a GAN-based approach that injects camera traces into synthetic images.  Given a GAN-generated image, we insert the traces of a specific camera model into it and deceive state-of-the-art detectors into believing the image was acquired by that model. Likewise, we deceive independent detectors of synthetic GAN images into believing the image is real. Experiments prove the effectiveness of the proposed method in a wide array of conditions. Moreover, no prior information on the attacked detectors is needed, but only sample images from the target camera.

\end{abstract}

\section{Introduction}

\begin{figure*}[t!]
    \centering
    \includegraphics[width=0.9\linewidth]{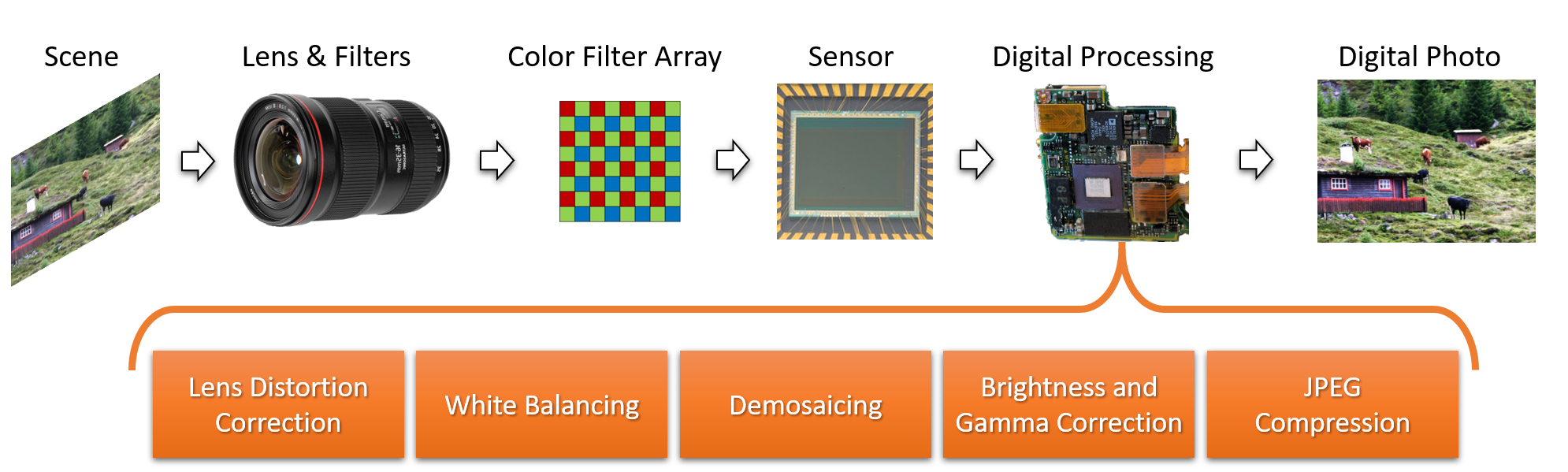}
    \vspace{-5pt}
    \caption{A digital image of a scene contains camera-related traces of the image formation process that could act as a fingerprint of a camera model. The used lenses and filters, the sensor and the manufacturer-specific digital processing pipelines result in unique patterns. These patterns can be used to identify camera models.}
    \label{fig:camera_pipeline}
\end{figure*}

There have been astonishing advances in synthetic media generation
in the last few years, thanks to deep learning,
and in particular to Generative Adversarial Networks (GANs).
This technology enabled significant improvement in the level of realism of generated data,
 increasing both resolution and quality \cite{Wang2018}.
Nowadays, powerful methods exist for generating an image from scratch \cite{Karras2018,Brock2019,karras2020analyzing},
and for changing its style \cite{Zhu2017,Karras2019, karras2020analyzing} or only some specific attributes \cite{Choi2018}.
These methods are very effective, especially on faces, and allow one to easily change the expression of a person \cite{Thies2016,Qian2019} or to modify its identity through face swapping \cite{Natsume2018,Nirkin2019}.
This manipulated visual content can be used to build more effective fake news.
In fact, it has been estimated that the average number of reposts for news containing an image is $11$ times larger than for those without images \cite{Jin2017}.
This raises serious concerns about the trustworthiness of digital content,
as testified by the growing attention to the deepfake phenomenon.
The research community has responded to this threat by developing a number of forensic detectors \cite{Verdoliva2020review}.
Some of them exploit high-level artifacts, like asymmetries in the color of the eyes, or anomalies arising from an imprecise estimation of the underlying geometry \cite{Matern2019,Yang2019}.
However, technology improves so fast that these visual artifacts will soon disappear.
Other approaches rely on the fact that any acquisition device leaves distinctive traces on each captured image \cite{Chen2008}, because of its hardware, or its signal processing suite.
They allow associating a media with its acquisition device at various levels,
from the type of source, to its brand/model, to the individual device \cite{Kirchner2015}.
A major impulse to this field has been given by the seminal work of Luk{\`{a}}s et al. \cite{Lukas2006},
where it has been shown that reliable device identification is possible based on the camera photo-response non-uniformity (PRNU) pattern.
This pattern is due to tiny imperfections in the silicon wafer used to manufacture the imaging sensor and can be considered as a type of device fingerprint.
Beyond extracting fingerprints that contain device-related traces, it is also possible to recover camera model fingerprints \cite{Cozzolino2020}.
These are related to the internal digital acquisition pipeline,
including operations like demosaicing, color balancing, and compression,
whose details differ according to the brand and specific model of the camera \cite{Kirchner2015} (See Fig.\ref{fig:camera_pipeline}).
Such differences help attribute images to their source camera, but can also be used to better highlight anomalies caused by image manipulations \cite{Ferrara2012,Agarwal2017}.
In fact, the absence of such traces, or their modification, is a strong clue that the image is synthetic or has been manipulated in some way \cite{Lyu2014,Cozzolino2020}.
Detection algorithms, however, must confront with the capacity of an adversary to fool them.
This applies to any type of classifier, and is also very well known in forensics,
where many counter-forensics methods have been proposed in the literature \cite{Barni2018}.
Indeed, forensics and counter-forensics go hand in hand, a competition that contributes to improving the level of digital integrity over time.
In this work, we propose a method to synthesize traces of cameras using a generative approach that is agnostic to the detector (i.e., not just targeted adversarial noise).
We achieve this by training a conditional generator to jointly fool an adversarial discriminator network as well as a camera embedding network.
To this end, the proposed method injects the distinctive traces of a target camera model in synthetic images, while reducing the original generation traces themselves,
leading all tested classifiers to attribute such images to the target camera ('targeted attack').
To the best of our knowledge, this is the first work that inserts real camera signatures into synthetic imagery to fool unknown camera identifier.
Indeed, previous work on fooling camera model identification, based on adversarial noise \cite{Guera2017,Marra2018vulnerability} or pattern injection \cite{Kirchner2009,Chen2019}, always considered only real images.
In addition, when generating the attack, assume advance knowledge of the attacked CNN-based classifier or, at least, all camera models used to train them \cite{Guera2017,Marra2018vulnerability,Chen2019}.
In contrast, we work in an open set scenario, and require exclusively a suitable number of images coming from the target camera model (in principle, we do not even need to know the camera model itself).
There is also some recent work that specifically fool GAN detectors by adding adversarial noise \cite{Carlini2020evading} or by removing low-level traces in GAN images \cite{Neves2020GANprintR}. However, the scenario is different from ours, since our aim is to make a synthetic image appear as a real one so as to fool not only a GAN detector, but also a camera model identification method.
Hence, our main contributions are the following:
\begin{itemize}
\item   we devise a GAN-based approach that inserts real camera traces in synthetic images;
\item   we carry out targeted attacks against CNN-based camera model detectors and show that they are easily fooled by our approach;
\item  we carry out attacks to GAN detectors without re-training our model and show that they can be easily fooled by our approach.
\end{itemize}

\section{Related work}

\paragraph{Adversarial attacks to camera identification.}
Adversarial attacks are conceived to fool a classifier by adding imperceptible perturbations \cite{Goodfellow2015}.
\cite{Guera2017} and \cite{Marra2018vulnerability} investigate on the vulnerability of deep learning based camera model classifiers in a white box setting.
They show that the attack is effective only when the image is uncompressed, 
while in realistic forensic applications the perturbation noise is hidden by the compression artifacts. 
The analyses also highlight the difficulty of transferring such attacks in the context of camera model identification. 
This is part of a more general behavior.
Contrary to what happens in many computer vision applications \cite{Goodfellow2015,Liu2017},
in forensics applications, where detectors rely on tiny variations of the data, 
transferability for non-targeted attacks is not achieved easily \cite{Gragnaniello2018,Barni2018a}.
A different perspective is followed in \cite{Chen2020camera} where, instead of introducing noise, 
camera traces are deleted from an image to prevent correct identification. 

\paragraph{Adversarial attacks to GAN image detection.}
Synthetic images do not contain camera-related traces.
However, they do contain hidden traces related to the pipeline used to generate them \cite{Marra2019, Zhang2019}.
These traces are implicitly or explicitly used to distinguish synthetic images from real ones by several CNN-based architectures \cite{Gragnaniello2021gan}. 

Some recent works proposed different ways to attack these detectors. In \cite{Carlini2020evading}
it has been investigated the robustness of such classifiers to adversarial attacks both in a white-box and in a black-box scenario. Experiments show that imperceptible perturbations can cause misclassification in both scenarios. A different perspective is pursued in \cite{Neves2020GANprintR}, where instead of adding noise, the specific fingerprints that characterize GAN images are removed through an autoencoder-based strategy.
While these papers are related to our work, there is a main difference since our objective is twofold.
In fact, our strategy is able to fool a GAN detector, but at the same time it can also fool a camera model identification method. In fact, we do not only reduce GAN traces, but we introduce the typical low-level features present in a target camera. This additional feature can be used in a forensic scenario to perform targeted attacks.

\paragraph{Adversarial attacks using generative networks.}
In the literature, several papers have already addressed the problem of using generative networks to create adversarial examples.
In \cite{Song2018} and \cite{Wang2019} the adversarial examples are generated from scratch, with no further constraint. Other papers instead are more relevant to our scenario and use a generative approach to slightly modify an existing image \cite{Hu2017,Poursaeed2018,Xiao2018,Chen2019,Deb2019}. 
In \cite{Poursaeed2018} the classifier under attack is supposed to be perfectly known (white-box scenario) and its gradient is used to train the generator of adversarial samples. In \cite{Hu2017}, instead, the attacker is only allowed to query the classifier and observe the predicted labels (black-box scenario). With this information, a substitute classifier is trained and its gradient is used to train the generator.
The strategies of \cite{Poursaeed2018} for white-box scenarios and of \cite{Hu2017} for black-box scenarios are adapted by Chen et al. \cite{Chen2019} for the specific problem to fool a camera model classifier.
In \cite{Xiao2018}, instead, these approaches are further extended by including a discriminator network to distinguish between original and attacked images, pushing the generator to improve the fidelity of the attacked image to the original. To adapt to a face recognition scenario, where the number of classes is not fixed in advance, in \cite{Deb2019} the target classifier is replaced with a face matcher based on the cosine similarity in the embedding space.

Unlike previous works, our proposal does not require knowledge of the classifier under attack \cite{Poursaeed2018,Xiao2018,Chen2019}, and not even of the labels output by the network in response to selected queries \cite{Hu2017,Xiao2018,Chen2019}. Moreover, although the use of an embedder was already proposed in \cite{Deb2019},  we use a less restrictive loss in the embedding space. In fact, while in \cite{Deb2019} the distance of the generated example is minimized with respect to a random sample of the target class, we minimize distances to a representative anchor vector, and focus mainly on critical outliers with respect to this anchor. In addition, our discriminator does not aim to preserve perceptual quality \cite{Xiao2018,Deb2019}, but to improve the generator’s ability to introduce traces of the target camera model, and to remove peculiar traces of synthetic images.
The objective is thus similar to \cite{Chen2019}; however,
in our work we inject camera model traces in synthetic images and not real ones.
More importantly, our strategy does not need to include the camera model classifier in the generation process as done in \cite{Chen2019},
but requires only images coming from the target distribution.

\section{Reference Scenario}
\label{sec:scenario}

In this section, we describe in more detail our reference scenario.
We want to show that inconsistencies in camera traces in forensics cannot be reliable even when the attacker has very little knowledge about the classifier. 

\vspace{-0.25cm}
\paragraph{Targeted attack:}
Our goal is to make a synthetic image appear as if it was taken by a specific real camera by inserting peculiar traces of the latter.
Hence, our attack is targeted, and our aim is to fool CNN-based camera model identification detectors
that will recognize the generated image as if it was taken by the target camera model.

\vspace{-0.25cm}
\paragraph{Available knowledge:}
We assume that the attacker has no knowledge about the specific classifier and cannot make queries to it.
We only suppose that the training images are drawn from the same distribution, that is, generated by the targeted camera model.
However, no knowledge is given on the other camera models on which the classifier is trained.

\vspace{-0.25cm}
\paragraph{Visual imperceptibility:}
We require that the attacked images look realistic and do not present visible artifacts.
The generation of realistic synthetic images is out of the scope of this work, and we simply assume they are available.
However, we require that the attack does not change the image content and introduces a perceptually acceptable distortion.

\vspace{-0.25cm}
\paragraph{Processing pipeline:}
Images are JPEG compressed after being modified to simulate a realistic scenario.
This is very important since fake content is especially harmful when it is uploaded on the internet and shared with a malicious goal to propagate false information.
\newcommand{\smean}[2]{\mathbb{E}_{#1} \left[ #2 \right] }

\newcommand{\R}{\mathcal{R}_M}
\renewcommand{\S}{\mathcal{S}}
\section{Proposed Method}
\label{sec:method}

Let
$\mathcal{R}_M$ be the set of real images generated by the target camera model $M$, and
$\mathcal{S}$ a set of synthetic images generated by some software tools.
We aim to process the images in $\mathcal{S}$ so as to make them forensically indistinguishable\footnote{Of course,
the synthetic images should also be visually realistic, but we do not address this problem, here, and assume the generation tool produces already visually plausible images.} 
from images in $\mathcal{R}_M$.
That is, we want to find a transformation $T_M(\cdot)$ such that, with high probability,
\[
    F_i(T_M(x)) = M, \hspace{6mm} x \in \mathcal{S}, \;\; i=1,\ldots,N
\]
with $F_i$'s the available camera model classifiers.
Since state-of-the-art forensic classifiers focus on the traces left on all acquired images by the model-specific processing suite, 
such traces must be injected into the original image.
We pursue this goal by training a convolutional neural network on images drawn from $\mathcal{R}_M$,
such that eventually, for each $x \in \mathcal{S}$, $y=T_M(x)$ looks to classifiers as belonging to $\mathcal{R}_M$.
Meanwhile, to remain undetected,
the attack is required not to modify the semantic content of the image, 
and hence to minimize some suitable measure of distortion $d(x,y)$ between original and modified images.

\newcommand{\graphcaption}{Our architecture is based on three components: a classical GAN setup, using a generator as well as a discriminator, and an embedding network. By applying a content loss between synthetic images as well as spoofed images, we ensure that our generator keeps the image content intact. The pre-trained embedder checks if we only generate specific camera traces. In comparison to regular GAN methods, we use all three involved images for our discriminator, i.e., synthetic, spoofed and camera typical images, and adjust the loss correspondingly.}

\begin{figure*}
    \centering
    \includegraphics[width=0.9\textwidth]{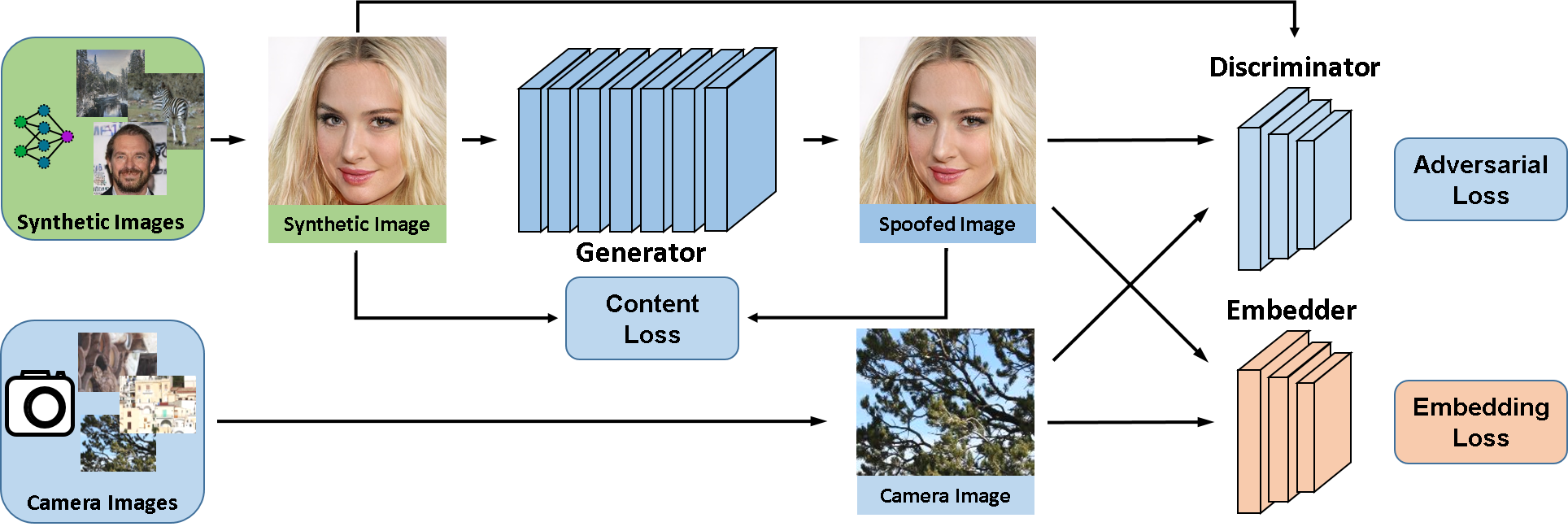}
    \vspace{-0.2cm}
    \caption{\graphcaption}
    \label{fig:generator_losses}
\end{figure*}

\subsection{Architecture}
Our goal is to be agnostic to any signature-based camera model identifier; i.e., we want more than inserting adversarial noise to fool a specific identifier.
To this end, we train a CNN through a reformulated GAN schema.
Specifically, our proposal involves the use of three networks: a generator, a discriminator and an embedder.
These are described in the following, and depicted in Fig.\ref{fig:generator_losses}.

\paragraph{Generator:}
The Generator, $G(\cdot)=T_M(\cdot)$, 
has the goal of introducing traces of a specific camera model $M$ into the input image, while preserving the semantic content.
We adopt an architecture formed by $7$ convolutional layers. 
The output of the last layer is summed to the input image and then a hyperbolic tangent is applied to limit the values of the output in the same range used for the input images, the range is equal to $[-1,1]$.
Note that, before entering the generator, the input image is Gaussian filtered with $\sigma=0.4$ to remove possible high-frequency imperfections, such as checkerboard artifacts.

\paragraph{Discriminator:}
In conventional GANs, the discriminator, $D(\cdot)$, is required to distinguish real from generated images, 
with the aim of pushing the generator to improve over time.
Accordingly, we could ask the discriminator to separate the sets 
$\R$, real images of the target camera model, and $G(\S)$, modified synthetic images.
Instead, we train it to tell apart $\R$ from both $G(\S)$ and $\S$, the set of original synthetic images.
By doing so, the generator is encouraged not only to introduce traces of the target camera model, 
but also to remove traces peculiar of synthetic images.
We use a patch-based discriminator \cite{Isola2017} realized as a fully-convolutional network with a fixed first layer. The fixed layer takes as input the RGB image and returns 9 channels, 
which are the original RGB components and their third-order horizontal and vertical derivatives.
In practice, this first layer extracts the image residuals, which highlight discriminating camera traces, 
speeding-up the network convergence \cite{Tuama2016}.

\paragraph{Embedder:}
In addition to a generator and discriminator, we use an embedder, $E(\cdot)$. 
This network is pre-trained off-line, using images drawn from many camera models, to extract a compact 512-dimensional feature vector.
It is trained using a triplet-loss \cite{Schroff2015} with the goal of obtaining the same feature vector 
for all images acquired by the same camera model.
Therefore, it provides a compact, model-specific, representation of the image, independent of the current data.
For the embedder, we use the same fixed first layer used for the discriminator.

\subsection{Loss function of the generator}
The goal of our scheme is to train an effective generator.
To this end, we define its objective function as the sum of three losses,
\begin{equation}
    \mathcal{L}_{G} = \mathcal{L}_{CNT} + \lambda_E \mathcal{L}_{EMB} + \lambda_A \mathcal{L}_{ADV}
    \label{equ:Lg}
\end{equation}
each of which drives the generator towards a specific goal:
preserving the scene content ($\mathcal{L}_{CNT}$), 
fooling the embedder ($\mathcal{L}_{EMB}$) and, 
fooling the discriminator ($\mathcal{L}_{ADV}$).
These losses are detailed in the following.

\paragraph{Scene content preserving loss:}
To achieve the first goal, we use a combination of an objective distortion measure as well as a perceptual loss between input and the output of the generator:
\begin{equation}
    \mathcal{L}_{CNT} = \mathcal{L}_{REC} + \lambda_p \mathcal{L}_{PER}
\end{equation}
As distortion measure we use an L1 distance between the two images, $\mathcal{L}_{REC} = \smean{x \sim \S}{ \|x - G(x)\|_1}$.
Following \cite{Johnson2016}, we define the perceptual loss $\mathcal{L}_{PER}$ as the sum of the L1 distances 
between the feature maps extracted by the VGG-19 network trained on ImageNet.

\paragraph{Embedder Loss:}
The second goal of the generator is to fool the embedder,
namely, to ensure that feature vectors extracted from the transformed images are indistinguishable from those of real images of the target model.
To this end, by averaging the feature vectors of real images, we first compute an anchor vector, 
$e_M = \smean{z \sim \R}{E(z)}$, representing the camera model in the embedding space.
Then we should aim at minimizing the distance, $d(x)=\| E(G(x))-e_M\|_1$, 
between feature vectors extracted from transformed images and this anchor vector. 
However, from the literature on triplet loss, it is well known that better results are obtained by comparing distances \cite{Schroff2015}.
Therefore, we first define a reference distance $d_{ref} = \smean{z \sim \R}{\| E(z)-e_M) \|_1 }$,
and then define the loss to minimize as 
\begin{equation}
    \mathcal{L}_{DST} = \smean{x \sim \S }{ \;|d(x) - d_{ref} + m|_+ }
\end{equation}
where $|x|_+ = x$ for $x>0$ and $0$ otherwise, and $m$ is the margin of the triplet-loss fixed to 0.01 in our experiments.
To further help the generator to fool the embedder, we include also a feature matching loss, $\mathcal{L}_{FM}$, 
proposed in literature \cite{Wang2018} to stabilize the training of GANs,
which we compute based on the feature maps extracted by the embedder of the real and the transformed images.
The final embedding loss is then
\begin{equation}
    \mathcal{L}_{EMB} = \mathcal{L}_{DST} + \lambda_f \mathcal{L}_{FM}
\end{equation}

\paragraph{Adversarial Loss:}
The discriminator, trained in parallel with the generator, 
should output values close to $1$ for real images, $x \in \R$, and close to zero for synthetic images, $x \in S$, both before and after being modified.
Accordingly, it relies on a modified binary cross-entropy loss:
\begin{equation}
\begin{split}
    \mathcal{L}_{D} =  - &              \smean{x \sim \R} { \log D(x) } + \\
                       - & \tfrac{1}{2} \smean{x \sim \S }{ \log (1-D(G(x))) + \log (1-D(x))}
\end{split}
\end{equation}
which pools original and modified synthetic images. 

A major goal of our generator is to modify the synthetic images to fool the discriminator, i.e., to make the discriminator believe they come from the target camera model.
Therefore, for the last term of Eq.(\ref{equ:Lg}) we adopt the standard adversarial loss based on the binary cross-entropy:
\begin{equation}
    \mathcal{L}_{ADV} = - \smean{x \sim \S }{ \log D(G(x)) }
\end{equation}
which is minimized when $D(G(x)) \to 1$, that is, synthetic images are classified as real after being modified.
Thus, generator and discriminator concur to modify the synthetic images 
to be similar to images of the target camera model but also different from the original synthetic images.

\newcommand{\ru}{\rule{0mm}{3mm}}

\section{Results}
\label{sec:results}

In this section we present the results of our experiments.
Specifically, we evaluate the proposed method in terms of its ability to 
(a) deceive detectors of GAN-generated images into believing they are dealing with real images (see Sec.~\ref{sec:gandec}) and,
(b) deceive camera model identifiers into recognizing the modified images as acquired by the target camera model (see Sec.~\ref{sec:camdec}).
Special attention will be devoted to testing robustness to off-training GANs.
To this end, experiments will be carried out on images generated by GAN architectures never seen in the training phase.
We consider a dataset of real images acquired by different camera models and a dataset of synthetic images generated by various GAN architectures.
For the real images, we use the 10 camera models adopted in the IEEE Forensic Camera Model Identification Challenge,
$400$ images per model are used for training, and $50$  for testing.
From the test set we sample $25000$-patches to test the camera model identifiers.
For the synthetic images, seven GAN architectures are considered: StarGAN \cite{Choi2018}, CycleGAN \cite{Zhu2017}, ProGAN \cite{Karras2018}, StyleGAN \cite{Karras2019}, RelGAN \cite{Wu2019relgan}, bigGAN \cite{Brock2018bigGAN}, and StyleGAN2 \cite{karras2020analyzing}.
For each of the first five architectures, we take $20000$ images for training and $2000$ for testing. 
In addition, $2000$ bigGAN images and $2000$ StyleGAN2 images are only used for test, but not in training, 
in order to evaluate generalization.
All experiments are carried out on $256\times256$-pixel patches.
For both real images and high-resolution GAN images (generated by ProGAN, StyleGAN and StyleGAN2) patches are extracted at random locations from the whole image.
For each of the $10$ camera models, a different generator-discriminator couple is trained, using only real images of the selected model.
We use ADAM optimizers with a batch size of $10$ and $30$ patches, respectively.
For both networks, the learning rate is set to $10^{-4}$, and the ADAM moments to $0.5$ and, $0.999$.
Through preliminary experiments, the loss weights are set to $\lambda_E=1$, $\lambda_A=0.1$, $\lambda_p=0.001$, and $\lambda_f=0.01$.
Training stops after $50K$ iterations.
The embedder is trained in advance using an external dataset of $600$ camera models and a total of $20394$ images publicly available on \url{dpreviewer.com}.
We also adopt ADAM for the embedder, with a batch of $80$ patches, a learning rate of $10^{-4}$ and default moments.
It is worth underlining that this dataset is different from the one used to train camera model identification methods, hence our approach does not assume a prior knowledge on the camera models used for the training step of the analyzed camera identification algorithms.

\subsection{Fooling camera model ID detectors}
\label{sec:camdec}

In this section we analyze the ability of our method to deceive camera model identifiers.
To this end, we consider four CNN-based target classifiers.
The first two architectures, Tuama2016~\cite{Tuama2016}, and Bondi2017~\cite{Bondi2017}, have been proposed specifically for camera model identification.
Moreover, in view of the results of the  IEEE Forensic Camera Model Identification Challenge 
hosted on the Kaggle platform\footnote{\url{https://www.kaggle.com/c/sp-society-camera-model-identification/}}
we also consider two deep general-purpose architectures, Xception~\cite{Chollet2017} and InceptionV3~\cite{Szegedy2016}, trained to work as camera model classifiers.
The performance of all these classifiers on our 10-models test set, in the absence of any attack, is shown in Tab.~\ref{tab:camera_model}.
Although all methods perform well, the very deep networks clearly outperform the other classifiers.
In general, most successful solutions are based on very deep networks \cite{Szegedy2016,Huang2017,Chollet2017}, that perform well even when data are subjected to
common post-processing operations, like re-compression.

\begin{table}[t!]
{\footnotesize
\centering
\begin{tabular}{|c|c|c|c|} \hline
\ru  Tuama2016 & Bondi2017 & Xception & InceptionV3   \\ \hline\hline
\ru  74.1 & 87.1 & 97.0 & 95.4    \\ \hline
\end{tabular}
\vspace{-5pt}
\caption{Accuracy (\%) of camera model identification evaluated on 25000 patches of dimension equal to $256\times256$ pixels.}
\label{tab:camera_model}
}
\end{table}

\subsubsection{Ablation study}
\label{sec:ablation}

To show the importance of the embedder and the discriminator in the training scheme, we compare two variants.
In the first variant, \textit{only-embed.}, we remove the discriminator by setting $\lambda_D=0$ for the loss of the generator (Equation 1 of the main paper).
While in the second variant, \textit{only-disc.}, we remove the embedder by setting $\lambda_E=0$ for the loss of the generator.
The other hyperparameters are not modified.
Performance is measured in terms of Successful Attack Rate (SAR), the fraction of modified synthetic images that are classified as acquired by the target model.
Tab.~\ref{tab:results_cam_variants} shows the performance of the variants to deceive four camera-model classifiers.
As can be seen this supports our choices.
The \textit{only-embed.} variant obtains the worst results with the maximum SAR of $48.13\%$ for Tuama2016 and a SAR lower than $15\%$ for the other classifiers, while the \textit{only-discr.} variant performs worse for all the four camera-model classifiers.

\begin{figure}
\begin{center}
\includegraphics[width=0.9\linewidth, page=1, trim=0 50 0 0]{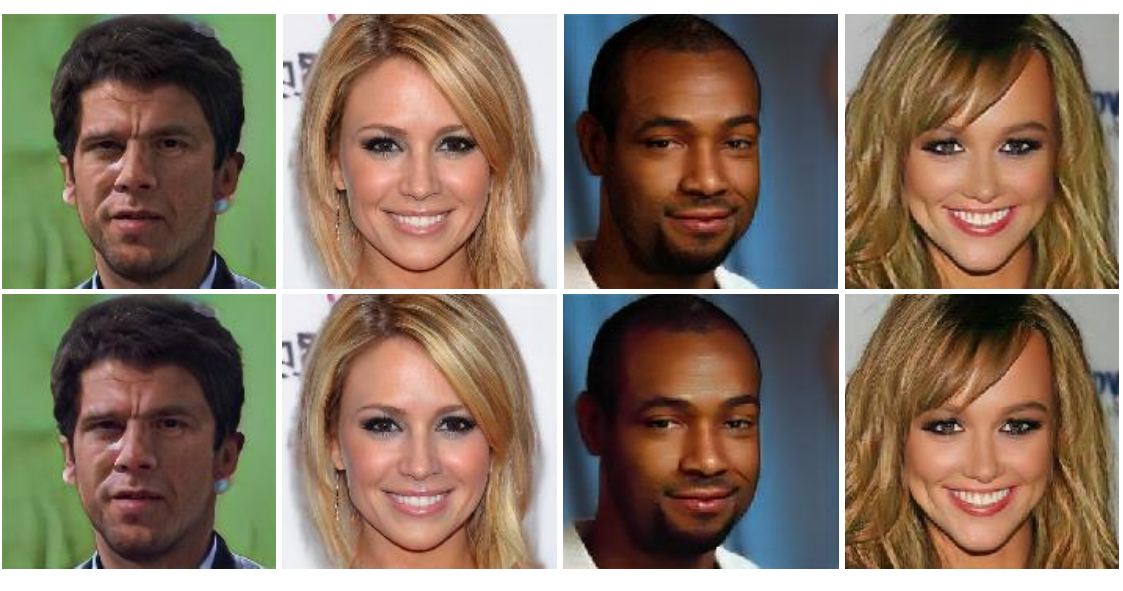}
\end{center}
   \caption{First line: original images from StyleGAN \cite{Karras2019}  and RelGAN \cite{Wu2019relgan}. Second line: attacked images with our method.}
\label{fig:examples}
\end{figure}

Finally, in Fig. \ref{fig:examples} we show some examples of attacked images, where no clear visual artifacts can be spotted.

\begin{table}[t!]
{\footnotesize
\centering
\begin{tabular}{|c||c||r|r|r|r||c|r|r|r|r||}  \cline{3-6}
   \multicolumn{2}{c|}{\ru~ }   & \multicolumn{4}{c||}{Model Classifiers}              \\ \hline    
\ru                     &   & Tuama & Bondi & Xcep. & Incep.  \\ 
\ru  Method                     &  PSNR & 2016 & 2017& & V3  \\ \hline \hline 
\ru Proposal                    & 31.7 &    \textbf{53.3}~ &   \textbf{67.6}~ &  \textbf{81.4}~ &     \textbf{71.7}~ \\ \hline 
\ru \textit{only-embed.}      & 49.7 &    48.1~ &    6.76~ &  10.8~ &     13.2~ \\ \hline  
\ru \textit{only-discr.} & 31.2 &    47.9~ &   63.8~ &  66.7~ &     70.4~ \\ \hline  
\end{tabular}
\vspace{-5pt}
\caption{Averaged results on 50000 images attacked using 5 different camera models. Performance is measured in terms of Successful Attack Rate (SAR).}
\label{tab:results_cam_variants}
}
\end{table}

\subsubsection{Comparison with state-of-the-art}
\label{sec:comparison}
We will compare the results of our proposal with four baseline methods for the generation of adversarial attacks:
Projected Gradient Descent (PGD)\footnote{\url{https://github.com/BorealisAI/advertorch}} \cite{Madry2017},
Translation-Invariant Momentum Iterative Fast Gradient Sign Method (TI-MI-FGSM) \cite{Dong2019},
Generative Adversarial Perturbation (GAP)\footnote{\url{https://github.com/OmidPoursaeed/Generative_Adversarial_Perturbations}} \cite{Poursaeed2018}, and
Generative Adversarial Attack against Camera Identification (Adv-Cam-Id) \cite{Chen2019}.
Only the latter method was specifically developed to attack camera model classifiers,
the other ones are more general and powerful approaches that insert adversarial noise to fool machine
learning based detectors. Adversarial noise in all these cases was added to
make the image recognized as if it was taken by a specific camera model.

All these methods perform white-box attacks, i.e., they are trained with reference to a known target classifier.
Here, we will consider both Bondi2017 and Xception as target classifiers.
However, to assess the transferability of the attack, we will evaluate performance also on all off-training classifiers.
To improve transferability, we also perform training an ensemble of classifiers, as proposed in \cite{Dong2019}; performance is evaluated on the fourth one.
For a fair comparison, we set the parameters of all methods in order to obtain a PSNR of $\approx$31dB.
We refer to the original papers for further details.
\newcommand{\mm}{\color{red}}
\begin{table}[t!]
{\footnotesize
\centering
\begin{tabular}{|c|l||c|r|r|r|r||c|r|r|r|r|} \cline{4-7}
  \cline{4-7}
   \multicolumn{3}{c|}{\ru~ }            & \multicolumn{4}{c|}{Model Classifiers} \\ \hline
\ru  &                 &   & Tuama  & Bondi & Xcep.  & Inc. \\ 
\ru      &  Method                &  PSNR & 2016~~  & 2017~~ &    & V3~~ \\ \hline\hline
\ru \multirow{4}{*}{\hspace{-5pt} \rotatebox{90}{Bondi2017~}}
    & PGD          & 30.98 &     1.0~ & {\mm  96.8~} &   0.0~ &      0.8~    \\ \cline{2-7}
\ru & TI-MI-FGSM   & 31.38 &    31.2~ & {\mm  31.7~} &   1.4~ &      2.0~    \\ \cline{2-7}
\ru & GAP          & 31.69 &    20.7~ & {\mm  67.4~} &   3.1~ &     22.4~    \\ \cline{2-7}
\ru & Adv-Cam-Id   & 30.18 &    24.1~ & {\mm  89.5~} &  54.5~ &     58.8~    \\ \hline\hline
\ru  \multirow{4}{*}{\hspace{-5pt} \rotatebox{90}{Xception~~~}}
    & PGD          & 33.52 &    24.7~ &    6.7~ & {\mm 98.2~} &      0.2~    \\ \cline{2-7}
\ru & TI-MI-FGSM   & 30.77 &    35.1~ &    3.2~ & {\mm 87.9~} &      6.5~    \\ \cline{2-7}
\ru & GAP          & 32.09 &     5.4~ &    0.6~ & {\mm 96.4~} &      5.5~    \\ \cline{2-7}
\ru & Adv-Cam-Id   & 29.69 &    37.6~ &   37.2~ & {\mm 97.5~} &     43.3~    \\ \hline\hline
\ru  \multirow{3}{*}{\hspace{-5pt} \rotatebox{90}{Ensemble}}
    & PGD          & 32.70 &    20.3~ &    9.2~ &   1.2~ &      0.5~    \\ \cline{2-7} 
\ru & TI-MI-FGSM   & 31.07 &    30.0~ &    3.2~ &   5.2~ &      5.5~    \\ \cline{2-7} 
\ru & GAP          & 31.97 &    25.5~ &    4.3~ &  12.1~ &      6.8~    \\ \hline\hline  
\ru & \textbf{SpoC (ours)} & 31.41 &    \textbf{55.6}~ &   \textbf{64.7}~ &  \textbf{73.4}~ &     \textbf{69.3}~    \\ \hline
\end{tabular}
\vspace{-5pt}
\caption{Averaging results on 50000 images attacked using 5 different camera models. Performance is measured in terms of a Successful Attack Rate (SAR). We compare with white-box attacks on Bondi2017 (first block) and Xception (second block) and hence discard values on such architecture for the analysis (red). We also compare with methods using an ensemble of classifiers (third block).}
\label{tab:results_cam}
}
\end{table}

\begin{table}[t!]
{\footnotesize
\centering
\begin{tabular}{|c|l||c|r|r|r|r|} 
  \cline{4-7}
   \multicolumn{3}{c|}{\ru~ }            & \multicolumn{4}{c|}{Model Classifiers} \\ \hline
\ru  &                 &   & Tuama  & Bondi & Xcep.  & Inc. \\ 
\ru      &  Method                &  PSNR & 2016~~  & 2017~~ &    & V3~~ \\ \hline\hline
\ru \multirow{4}{*}{\hspace{-5pt} \rotatebox{90}{Bondi2017~}}
    & PGD                   & 32.33 &  12.2~ & {\mm 97.5~} &   3.5~ &  13.5~   \\ \cline{2-7}
\ru & TI-MI-FGSM            & 31.29 &  34.1~ & {\mm 27.9~} &   2.1~ &   5.1~   \\ \cline{2-7}
\ru & GAP                   & 31.55 &  16.3~ & {\mm 59.4~} &   4.0~ &  22.5~   \\ \cline{2-7}
\ru & Adv-Cam-Id            & 24.63 &  23.5~ & {\mm 71.8~} &  33.6~ &  39.8~   \\ \hline\hline
\ru  \multirow{4}{*}{\hspace{-5pt} \rotatebox{90}{Xception~~~}}
    & PGD                   & 33.11 &  20.2~ &   5.6~ & {\mm 98.5~} &   0.2~   \\ \cline{2-7}
\ru & TI-MI-FGSM            & 30.56 &  37.1~ &   4.0~ & {\mm 93.3~} &  11.0~   \\ \cline{2-7}
\ru & GAP                   & 31.88 &   3.6~ &   1.2~ & {\mm 90.8~} &   2.5~   \\ \cline{2-7}
\ru & Adv-Cam-Id            & 28.79 &  34.7~ &  40.1~ & {\mm 98.0~} &  40.3~   \\ \hline\hline
\ru  \multirow{3}{*}{\hspace{-5pt} \rotatebox{90}{Ensemble}}
    & PGD                   & 32.39 &  18.4~ &   4.4~ &   1.6~ &   0.2~   \\ \cline{2-7}  
\ru & TI-MI-FGSM            & 30.93 &  35.0~ &   2.1~ &   6.4~ &   9.4~   \\ \cline{2-7}  
\ru & GAP                   & 31.90 &  24.8~ &   2.1~ &  20.4~ &   5.3~   \\ \hline\hline
\ru & \textbf{SpoC (ours)}  & 30.37 & \textbf{56.8~} & \textbf{48.4~} & \textbf{74.2~} & \textbf{66.0~}   \\ \hline
\end{tabular}
\vspace{-5pt}
\caption{Averaging results on 20000 images attacked using 5 different camera models and two GAN architectures (\cite{Brock2018bigGAN}, \cite{karras2020analyzing}) outside the training set. Performance is measured in terms of a Successful Attack Rate (SAR). We compare with white-box attacks on Bondi2017 (first block) and Xception (second block) and hence discard values on such architecture for the analysis (red). We also compare with methods using an ensemble of classifiers (third block).}
\label{tab:results_cam_out}
}
\end{table}
Results are computed on the central $256\times256$ crop of $10000$ synthetic images, and averaged over $5$ target models 
(Motorola DroidMaxx, Samsung GalaxyS4, Sony Nex7, iPhone 4s and, Motorola X).
All images are JPEG compressed using the quantization table of the target camera model.
This corresponds to a realistic setting where images are compressed before being spread over the web.
Note that in the absence of any attack, the average SAR is 10\% for our 10-class setting.
Results are reported in Tab.~\ref{tab:results_cam}.
Several considerations are in order.
First of all, it is clear that all white-box attacks are very effective when tested on the very same classifier used during training, Bondi2017 in the first block, Xception in the second one.
Such results are emphasized with red text.
However, there is no reason to expect such a scenario in practice, as the defender is free to use any classifier for camera identification.
Therefore, the most interesting results are those in black.
On off-training classifiers, only Adv-Cam-Id, among the baselines, provides a reasonably good performance, never exceeding  $60\%$ though.
Instead, SpoC performs quite well on all classifiers, with SAR going from about $56\%$ to $73\%$.
With respect to Adv-Cam-Id, the best baseline, it improves from $10\%$ to $30\%$.
Training on an ensemble of classifiers should ensure higher transferability, but the results of the third block do not seem especially encouraging, with an average SAR barely exceeding $10\%$.

Although, in Tab.~\ref{tab:results_cam}, we show results on $256\times256$ crops, our method can also be applied to higher resolution images (e.g., $1024\times 1024$) without re-training the network.
Indeed, if we apply our network to the StyleGAN images at resolution $1024\times1024$, we still achieve a good result with an attack success rate always above $48\%$.

\begin{table*}[!t]
	{\footnotesize
		\centering
		\begin{tabular}{l||c|c|c|c|c|c|c|c|c|c|} \cline{2-11}
			\ru               & \multicolumn{5}{c|}{in training} & \multicolumn{5}{c|}{out training} \\ \cline{2-11}
			\ru               & \multirow{2}{*}{Xception} & \multirow{2}{*}{~~~Spec~~~} & \multirow{2}{*}{ResNet50} & Patch     & \multirow{2}{*}{~~~FFD~~~} & \multirow{2}{*}{Xception} & \multirow{2}{*}{~~~Spec~~~} & \multirow{2}{*}{ResNet50}  & Patch     & \multirow{2}{*}{~~~FFD~~~} \\ 
			\ru           	  &          &            &          & Forensics &           &          &            &           & Forensics &           \\ \hline\hline
			\ru  before our attack    & 99.87 & 76.06 & 99.38 & 78.90 & 97.83    & 82.12 & 22.02 & 80.55 & 22.77 & 75.94\\ \hline
			\ru  after our attack & 44.28 & ~8.45 & 45.69 & ~3.65 & 27.19    & ~1.00 & ~0.96 & ~5.24 & ~0.11 &  4.74\\ \hline
		\end{tabular}
		\vspace{-5pt}
		\caption{True Positive Rate (TPR) for the GAN detectors before and after the proposed attack using GAN architectures considering both images inside (StarGAN \cite{Choi2018}, CycleGAN \cite{Zhu2017}, ProGAN \cite{Karras2018}, StyleGAN \cite{Karras2019}, and RelGAN \cite{Wu2019relgan}) and outside the training-set (bigGAN \cite{Brock2018bigGAN}, and StyleGAN2 \cite{karras2020analyzing}).}
		\label{tab:TPR_GAN_detectors}
	}
\end{table*}

\subsubsection{Generalization}
\label{sec:generalization}
Both the proposed technique and the reference techniques based on a generative network require a training phase.
To test their ability to generalize, we add a further experiment, on images generated by a GAN architecture not used in the training set.
In Table \ref{tab:results_cam_out} we show the results.

GAP shows about the similar performance, measured in terms of SAR and PSNR, on data generated by architectures outside and inside the training-set.
On the contrary Adv-Cam-Id presents a reduction of PSNR (until 5dB), while preserving a reasonably good SAR.
Finally, comparing the SpoC with respect to the other methods, it continues to have the better results, with SAR going from $48.4\%$ to $74.2\%$, with a minimal reduction in PSNR.

\subsection{Fooling GAN-image detectors}
\label{sec:gandec}
Here, we show that our architecture largely succeeds in removing the peculiar features of GAN images that make them distinguishable from real images. It is important to highlight that we do not re-train our model.
To this end, we challenge five CNN-based GAN-image detectors.
The first one is the spectrum-based (Spec) classifier proposed in \cite{Zhang2019}, 
which detects the frequency-domain peaks caused by the up-sampling steps of common GAN pipelines.
Moreover, we consider a two general-purpose deep networks, Xception \cite{Chollet2017} and ResNet50 \cite{He2016deep}, which proved to be an effective tool for GAN image detection and deepfakes video \cite{Marra2018gan, Roessler2019, Wang2019dec}.
We also include PatchForensics that is a fully-convolutional patch-based classifier proposed in \cite{Chai2020makes} and
FFD (Facial Forgery Detection)~\cite{Dang2020detecting}, a variant of Xception, that includes an attention-based layer, in order to focus on high-frequency details. 
All the detectors are trained on $10000$ synthetic images coming from 5 GAN architectures and $4000$ real images coming from 10 cameras. We used the augmentation strategy proposed in \cite{Wang2019dec}, that helped to increase the generalization ability of each detector.  
Testing is carried out on $256\times256$-pixel central crop of $10000$ synthetic images of seen GAN architectures and $4000$ synthetic images of two unseen GAN architectures.
In Tab.~\ref{tab:TPR_GAN_detectors} (left), we show the results of the experiment with aligned training and test.
The detectors have a high true positive rate (TPR), sometimes close to $100\%$, that is, they detect GAN images with near certainty. They have also good performance on real images with a false positive rate (FPR), not shown in the table, always less than $1\%$.
However, after modifying the synthetic images with our attack, the TPRs decrease to $3.65\%$ and, $45.69\%$.
In a second experiment, we work on the off-training images generated by the BigGAN and StyleGAN2 architecture.
Results are reported in Tab.~\ref{tab:TPR_GAN_detectors} (right).
Some detectors (Xception, ResNet50 and FFD) keeps detecting GAN images with high accuracy,
nonetheless, after modifying the images with our approach, the TPR reduces drastically less than $6\%$.
\section{Conclusion}

In this work, we proposed a GAN-based method to attack both forensic camera model identifiers and GAN detectors.
Our scheme allows to inject model-specific traces into the input image
such to deceive the classifier into believing the image was acquired by the desired target camera model.
Moreover, the method requires no prior knowledge on the attacked network, 
and works even on completely synthesized images.
Experimental results prove the effectiveness of the proposed method and, 
as a consequence, the weaknesses of current forensic detectors, 
calling for new approaches, less dependant on critical hypotheses, and more resilient to unforeseen attacks.

\section*{Acknowledgement}
 We gratefully acknowledge the support of this research by the AI Foundation, a TUM-IAS Hans Fischer Senior Fellowship, and Google Faculty Research Award. In addition, this material is based on research sponsored by the Defense Advanced Research Projects Agency (DARPA) and the Air Force Research Laboratory (AFRL) under agreement number FA8750-20-2-1004. The U.S. Government is authorized to reproduce and distribute reprints for Governmental purposes notwithstanding any copyright notation thereon.
 The views and conclusions contained herein are those of the authors and should not be interpreted as necessarily representing the official policies or endorsements, either expressed or implied, of DARPA and AFRL or the U.S. Government. This work is also supported by the PREMIER project, funded by the Italian Ministry of Education, University, and Research within the PRIN 2017 program.

\begin{appendix}
\section*{Appendix}
%
%
In this appendix, we report the details of the architectures used for Generator, Discriminator, and Embedder  (Sec.~\ref{sec:architectures}).
For reproducibility, we report the parameters of the used comparison methods used in the main paper (see Sec.~\ref{sec:comparisonw}).
Finally, we analyze the scenario where we attack images by varying the JPEG compression level.
In addition, we report the results obtained when we want to fool at the same time both a model classifier and a GAN detector (see Sec.~\ref{sec:further_exp}).

\begin{figure*}[!b]
    \centering
    \includegraphics[scale=0.35]{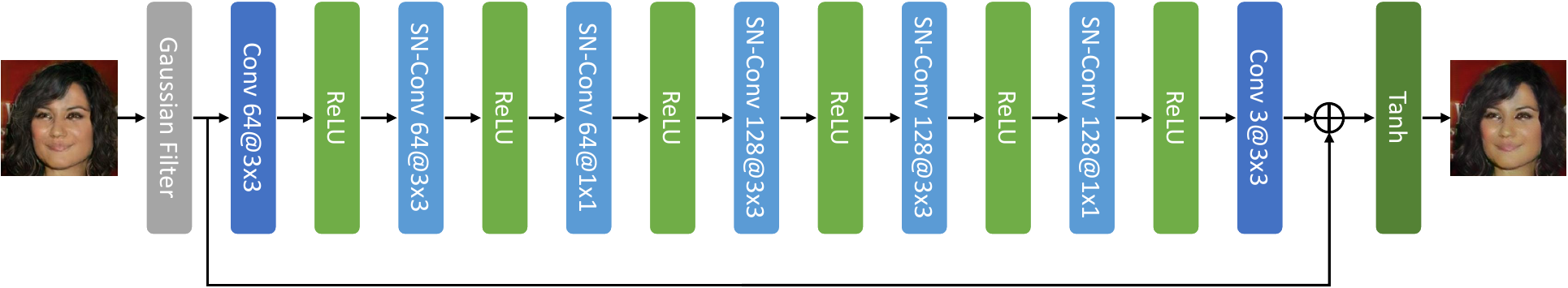}
    \caption{Generator architecture.}
    \label{fig:generator}
\end{figure*}

\begin{figure*}[!b]
    \centering
    \includegraphics[scale=0.35]{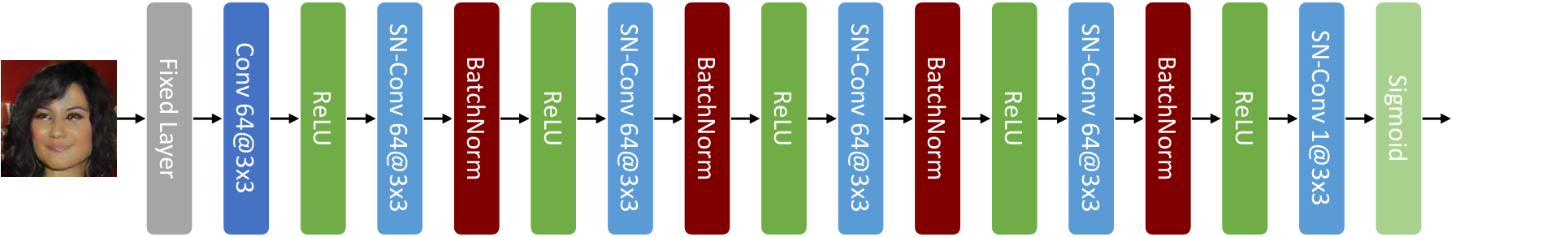}
    \caption{Discriminator architecture.}
    \label{fig:discriminator}
\end{figure*}

\section{Architectures}
\label{sec:architectures}
\paragraph{Generator} Our generator is composed of seven convolutional layers with a fixed stride equal to one (see Fig.\ref{fig:generator}).
The number of feature channels increases through the network from 64 to 128 after the first three convolutional layers and is set to the image channel size of three in the last layer.
We apply appropriate padding to keep the input image dimensions of 256$\times$256. In order to guide our adversarial training, we apply spectral normalization for our five middle layers as described in \cite{miyato2018spectral}.
We use ReLU as non-linearity for all layers besides the last.
After the input has been passed through our convolutional layers, we use a residual connection to add it to our output and squash the final result back to image space using a Tanh non-linearity.

\paragraph{Discriminator} As described in the main paper, the discriminator uses a fixed first layer to extract low level image features.
This input is fed into a convolutional layer with a kernel size of three.
Afterwards, we use four blocks of convolutional layers using a kernel size of three, spectral as well as mean-only batch normalization.
The number of feature channels is 64  for all these layers and we use ReLU as non-linearity.
The output is fed into a final convolutional layer with kernel size of three to reduce the number of features to one.
We use no padding for all convolutional layers in our discriminator.
The discriminator architecture is shown in Fig.\ref{fig:discriminator}.

\begin{figure}[t]
    \centering
    \includegraphics[scale=0.35]{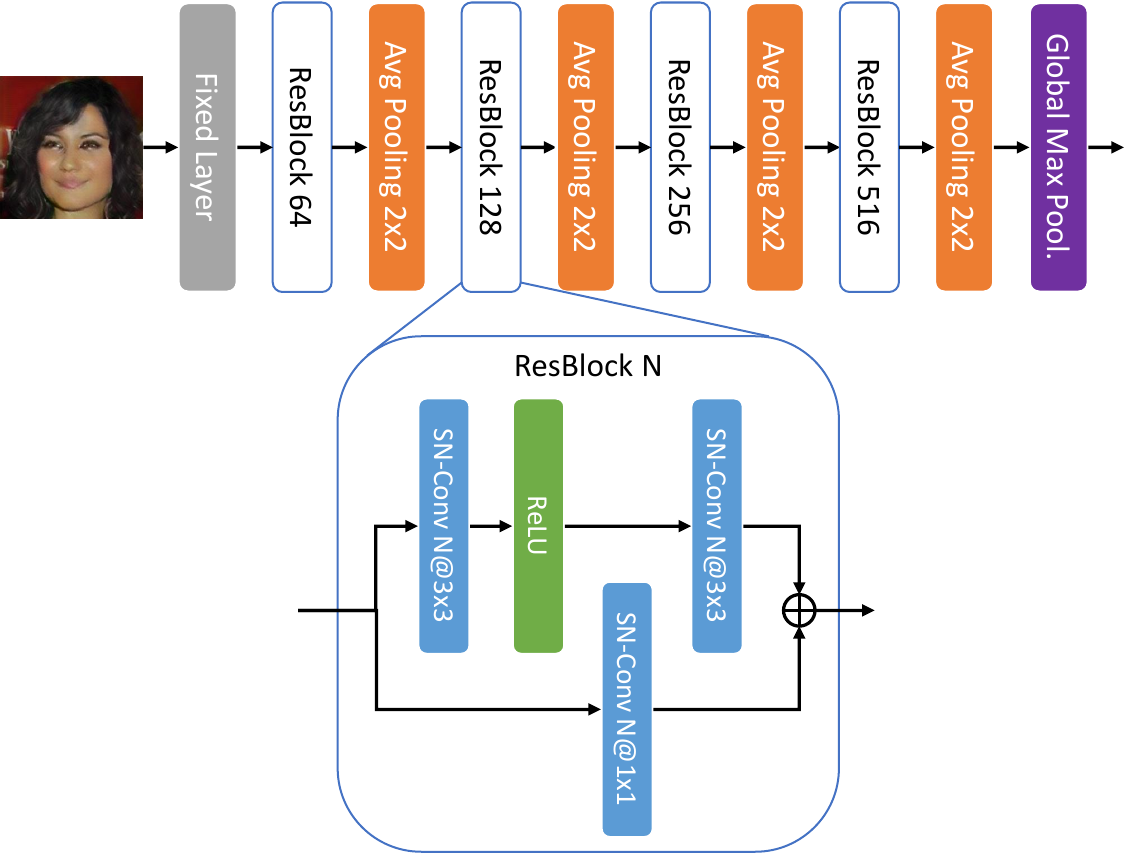}
    \caption{Embedder architecture.}
    \label{fig:embedder}
\end{figure}

\paragraph{Embedder} As shown in Fig.\ref{fig:embedder}, the main layer in our embedder is a residual block.
This block has two branches. In one branch, a convolution with kernel size one is applied, while in the other branch, we make use of two convolutions using a kernel size of three together with a ReLU non-linearity.
The outputs of the two branches are then summed up to obtain the final output tensor.
We adopt spectral normalization for all convolutional layers. 
As described in the paper, the input image of the embedder is first passed through our fixed layer to extract low level image features.
The output is passed through four residual blocks following each one by an average pooling of size two.
The number of feature channels linearly increases from 64 to 512.
The output is pooled to a single 512 dimensional tensor using a global max pooling.

\section{Comparison with state-of-the-art}
\label{sec:comparisonw}

In the main paper we compare our proposal with four techniques that generate adversarial attacks.
For all these techniques, we set the parameters in order to obtain a PSNR of about 31dB.
In the following, we give more details about these techniques.

\paragraph{PGD (Projected Gradient Descent attack) \cite{Madry2017}:}
PGD is an iterative attack method based on the evaluation of the
gradient of the loss function w.r.t the input image. At each iteration, the image is modified with the projection of the gradient into the space of allowed perturbations. For this method, we use a number of iterations equal to 40 and an epsilon for each attack iteration equal to $1.25$. 

\paragraph{TI-MI-FGSM (Translation-Invariant Momentum Iterative Fast Gradient Sign Method) \cite{Dong2019}:}
It is an iterative version of the Fast Gradient Sign Method with the use of a momentum term for the estimation of the gradient. Moreover, to improve the transferability of the attack, the gradient is computed considering a set of translated versions of the image. For this method, we use a number of iterations equal to 40, an epsilon for each attack iteration equal to $0.4$ and, an overall epsilon equal to 8.

\paragraph{GAP (Generative Adversarial Perturbation) \cite{Poursaeed2018}:}
It is a method where a generator network is trained in order to obtain a perturbation able to fool the classifier with a constraint on the maximum allowed perturbation.
The authors propose two variants, Universal Perturbation, and  Image-dependent Perturbation.
In the first case, the perturbation does not directly depend on the image to attack, while in the second case it depends on the image.
We compare the proposal with Image-dependent Perturbation that is a less restrictive hypothesis and more coherent with our scenario.
As proposed by the authors, for the architecture of the generator network, we use ResNet Generator that is defined in \cite{Johnson2016}.
In our experiments, the epsilon of the constraint is set equal to 8.

\paragraph{Adv-Cam-Id \cite{Chen2019}:} 
The white-box attack proposed in \cite{Chen2019} uses a generator network that provides a falsified version of the image.
The generator network is trained using two losses, one is relative to the capability to fool the classifier and
the other is the L1 distance between the original image and the falsified image. 
The generator architecture is composed of a first block, that emulates the color filter array, and seven convolutional layers.
For our experiments, we use the hyperparameters suggested by the authors and stop the training when the PSNR is greater than or equal to 31dB.

\begin{figure}[t]
	\centering
	\includegraphics[width=0.95\linewidth]{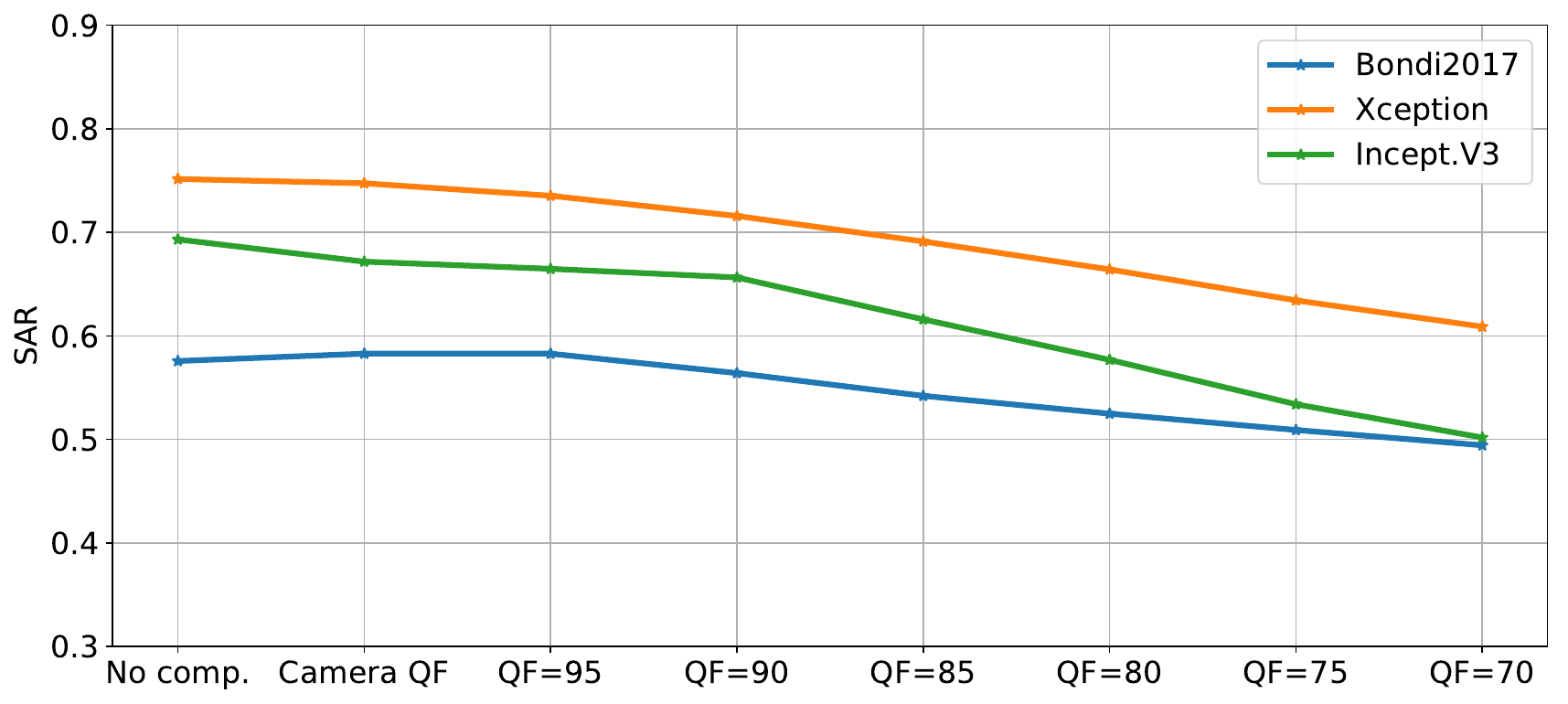}
	\caption{Attack success rate of our proposal by varying the compression level of the attacked images.}
	\label{fig:sar_quality}
\end{figure}

\begin{figure}[t]
	\centering
	\includegraphics[width=0.95\linewidth]{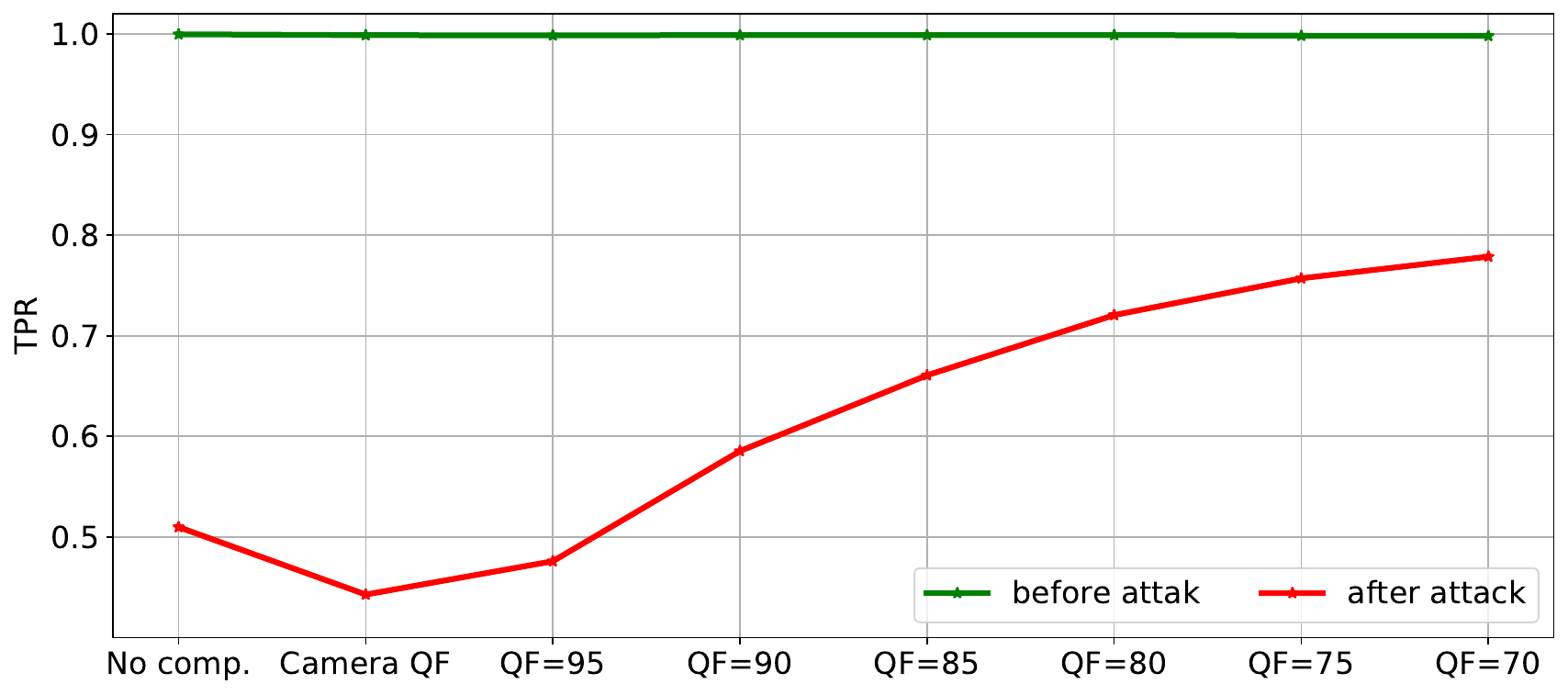}
	\caption{True Positive Rate of the network Xception by varying the compression level before and after the proposed attack.}
	\label{fig:tpr_quality}
\end{figure}

\section{Additional results}
\label{sec:further_exp}

In this section we analyze robustness to compression of our approach.
In Fig.\ref{fig:sar_quality} we show the attack success rate by varying the JPEG compression quality.
Camera model classifiers have been trained using images at different JPEG compression levels, so as to improve their performance also on compressed data.
In this analysis we exclude Tuama2016 because it achieves an accuracy below 50\% in this scenario.
Results are shown in Fig.\ref{fig:sar_quality} and show that the attack is still effective even on compressed images,
in particular the attack success rate still remains above 50\% also for images compressed at JPEG quality in the range $[70-75]$,
which is the quality level typically applied by a social networks when an image is uploaded.

We also carry out a similar analysis on GAN detectors. 
The trend by varying the JPEG compression is very much similar for all the GAN detectors,
hence in Fig.\ref{fig:tpr_quality} we only report the behavior of Xception by varying the compression level.
Performance of the detector are perfect before the attack and then reduce strongly after our attack, even on heavily compressed images.

Finally, we consider a scenario where the objective is to fool both the camera model classifier and the GAN detector at the same time. Results are shown in Tab.\ref{tab:SAR_both} and confirm that our approach is able to obtain satisfying results, especially when attacking deeper networks for camera model identification and GAN detectors that were not trained on that specific GAN images.

\begin{table*}[t!]
	{\footnotesize
		\centering
		\begin{tabular}{|c|l||c|c|c|c|c|c|c|c|c|c|} \cline{3-12}
			\multicolumn{2}{c|}{\ru}              & \multicolumn{5}{c|}{in training} & \multicolumn{5}{c|}{out training} \\ \cline{3-12}
			\multicolumn{2}{c|}{\ru}            & \multirow{2}{*}{Xception} & \multirow{2}{*}{~~~Spec~~~} & \multirow{2}{*}{ResNet50} & Patch     & \multirow{2}{*}{~~~FFD~~~} & \multirow{2}{*}{Xception} & \multirow{2}{*}{~~~Spec~~~} & \multirow{2}{*}{ResNet50}  & Patch     & \multirow{2}{*}{~~~FFD~~~} \\ 
			\multicolumn{2}{c|}{\ru}  	  &          &            &          & Forensics &           &          &            &           & Forensics &           \\ \hline\hline
			\ru \multirow{4}{*}{\hspace{-5pt} \rotatebox{90}{Model  Clas.}}  &Tuama2016   &  34.8  &  51.2  &  31.0  &  54.0  &  40.8  &  56.3  &  56.4  &  53.5  &  56.8  &  55.1  \\ \cline{2-12}
            \ru &Bondi2017   &  38.8  &  59.3  &  33.9  &  62.6  &  47.2  &  47.9  &  47.8  &  45.0  &  48.4  &  46.5  \\ \cline{2-12}
            \ru &Xception    &  43.8  &  66.3  &  37.9  &  71.1  &  54.6  &  73.6  &  73.4  &  69.8  &  74.2  &  72.5  \\ \cline{2-12}
            \ru &InceptionV3 &  40.6  &  63.1  &  38.8  &  67.7  &  51.0  &  65.3  &  65.3  &  61.8  &  66.0  &  63.5  \\ \hline
		\end{tabular}
		\vspace{-5pt}
		\caption{Successful  Attack  Rate  (SAR) to fool the Model classifier and the GAN detector at the same time considering both images inside (StarGAN \cite{Choi2018}, CycleGAN \cite{Zhu2017}, ProGAN \cite{Karras2018}, StyleGAN \cite{Karras2019}, and RelGAN \cite{Wu2019relgan}) and outside the training-set (bigGAN \cite{Brock2018bigGAN}, and StyleGAN2 \cite{karras2020analyzing}).
		\label{tab:SAR_both}
		}
	    }
\end{table*}

~~~~~~~~~~~~~~~~~~~~~~~~~~~~~~~~~~~~~~~~~
\end{appendix}

\balance
{\small
\bibliographystyle{ieee_fullname}
\bibliography{egbib}

\begin{thebibliography}{10}\itemsep=-1pt

\bibitem{Agarwal2017}
Shruti Agarwal and Hany Farid.
\newblock {Photo Forensics from JPEG Dimples}.
\newblock In {\em IEEE International Workshop on Information Forensics and
  Security}, 2017.

\bibitem{Barni2018a}
Mauro Barni, Kassem Kallas, Ehsan Nowroozi, and Benedetta Tondi.
\newblock {On the Transferability of Adversarial Examples against CNN-based
  Image Forensics}.
\newblock In {\em IEEE International Conference on Acoustics, Speech and Signal
  Processing (ICASSP)}, 2018.

\bibitem{Barni2018}
Mauro Barni, Matthew Stamm, and Benedetta Tondi.
\newblock {Adversarial multimedia forensics: Overview and challenges ahead}.
\newblock In {\em European Signal Processing Conference (EUSIPCO)}, pages
  962--966, 2018.

\bibitem{Bondi2017}
Luca Bondi, Luca Baroffio, David G{\"{u}}era, Paolo Bestagini, Edward Delp, and
  Stefano Tubaro.
\newblock First steps toward camera model identification with convolutional
  neural networks.
\newblock {\em IEEE Signal Processing Letters}, 24(3):259--263, 2017.

\bibitem{Brock2018bigGAN}
Andrew Brock, Jeff Donahue, and Karen Simonyan.
\newblock Large scale gan training for high fidelity natural image synthesis,
  2018.

\bibitem{Brock2019}
Andrew Brock, Jeff Donahue, and Karen Simonyan.
\newblock Large scale {GAN} training for high fidelity natural image synthesis.
\newblock In {\em International Conference on Learning Representations}, 2019.

\bibitem{Carlini2020evading}
Nicholas Carlini and Hany Farid.
\newblock Evading deepfake-image detectors with white- and black-box attacks.
\newblock In {\em IEEE/CVF Conference on Computer Vision and Pattern
  Recognition Workshops (CVPRW)}, pages 2804--2813, 2020.

\bibitem{Chai2020makes}
Lucy Chai, David Bau, Ser-Nam Lim, and Phillip Isola.
\newblock What makes fake images detectable? understanding properties that
  generalize.
\newblock In {\em European Conference on Computer Vision (ECCV)}, pages
  103--120, 2020.

\bibitem{Chen2020camera}
Chang Chen, Zhiwei Xiong, Xiaoming Liu, and Feng Wu.
\newblock Camera trace erasing.
\newblock In {\em IEEE/CVF Conference on Computer Vision and Pattern
  Recognition (CVPR)}, pages 2947--2956, 2020.

\bibitem{Chen2019}
Chen Chen, Xinwei Zhao, and Matthew~C. Stamm.
\newblock Generative adversarial attacks against deep-learning-based camera
  model identification.
\newblock {\em IEEE Trans. Inf. Forensics Security, in press}, October 2019.

\bibitem{Chen2008}
Mo Chen, Jessica Fridrich, Miroslav Goljan, and Jan Luk{\`{a}}{\v{s}}.
\newblock Determining image origin and integrity using sensor noise.
\newblock {\em IEEE Trans. Inf. Forensics Security}, 3:74--90, 2008.

\bibitem{Choi2018}
Yunjey Choi, Minje Choi, Munyoung Kim, Jung-Woo Ha, Sunghun Kim, and Jaegul
  Choo.
\newblock {StarGAN: Unified Generative Adversarial Networks for Multi-Domain
  Image-to-Image Translation}.
\newblock In {\em IEEE Conference on Computer Vision and Pattern Recognition
  (CVPR)}, 2018.

\bibitem{Chollet2017}
Fran{\c{c}}ois Chollet.
\newblock Xception: Deep learning with depthwise separable convolutions.
\newblock In {\em IEEE Conference on Computer Vision and Pattern Recognition
  (CVPR)}, 2017.

\bibitem{Cozzolino2020}
Davide Cozzolino and Luisa Verdoliva.
\newblock {Noiseprint: a CNN-based camera model fingerprint}.
\newblock {\em IEEE Trans. Inf. Forensics Security}, 15:144--159, 2020.

\bibitem{Dang2020detecting}
Hao Dang, Feng Liu, Joel Stehouwer, Xiaoming Liu, and Anil~K Jain.
\newblock On the detection of digital face manipulation.
\newblock In {\em IEEE conference on Computer Vision and Pattern Recognition
  (CVPR)}, pages 5781--5790, 2020.

\bibitem{Deb2019}
Debayan Deb, Jianbang Zhang, and Anil~K. Jain.
\newblock Advfaces: Adversarial face synthesis.
\newblock In {\em IEEE International Joint Conference on Biometrics (IJCB)},
  pages 1--10, 2020.

\bibitem{Dong2019}
Yinpeng Dong, Tianyu Pang, Hang Su, and Jun Zhu.
\newblock Evading defenses to transferable adversarial examples by
  translation-invariant attacks.
\newblock In {\em IEEE Conference on Computer Vision and Pattern Recognition
  (CVPR)}, 2019.

\bibitem{Ferrara2012}
Pasquale Ferrara, Tiziano Bianchi, Alessia~De Rosa, and Alessandro Piva.
\newblock {Image forgery localization via fine-grained analysis of CFA
  artifacts}.
\newblock {\em IEEE Trans. Inf. Forensics Security}, 7(5):1566--1577, 2012.

\bibitem{Goodfellow2015}
Ian Goodfellow, Jonathon Shlens, and Christian Szegedy.
\newblock Explaining and harnessing adversarial examples.
\newblock In {\em International Conference on Learning Representations}, 2015.

\bibitem{Gragnaniello2021gan}
Diego Gragnaniello, Davide Cozzolino, Francesco Marra, Giovanni Poggi, and
  Luisa Verdoliva.
\newblock {Are GAN generated images easy to detect? A critical analysis of the
  state-of-the-art}.
\newblock In {\em IEEE International Conference on Multimedia and Expo (ICME)},
  2021.

\bibitem{Gragnaniello2018}
Diego Gragnaniello, Francesco Marra, Giovanni Poggi, and Luisa Verdoliva.
\newblock {Analysis of adversarial attacks against CNN-based image forgery
  detectors}.
\newblock In {\em European Signal Processing Conference}, pages 384--389, 2018.

\bibitem{Guera2017}
David G{\"u}era, Yu Wang, Luca Bondi, Paolo Bestagini, Stefano Tubaro, and
  Edward Delp.
\newblock {A Counter-Forensic Method for CNN-Based Camera Model
  Identification}.
\newblock In {\em IEEE CVPR Workshops}, 2017.

\bibitem{He2016deep}
Kaiming He, Xiangyu Zhang, Shaoqing Ren, and Jian Sun.
\newblock Deep residual learning for image recognition.
\newblock In {\em IEEE Conference on Computer Vision and Pattern Recognition
  (CVPR)}, pages 770--778, 2016.

\bibitem{Hu2017}
Weiwei Hu and Ying Tan.
\newblock {Generating Adversarial Malware Examples for Black-Box Attacks Based
  on GAN}.
\newblock {\em arXiv preprint arXiv:1702.05983v1}, 2019.

\bibitem{Huang2017}
Gao Huang, Zhuang Liu, Laurens Van Der~Maaten, and Kilian~Q. Weinberger.
\newblock Densely connected convolutional networks.
\newblock In {\em IEEE Conference on Computer Vision and Pattern Recognition
  (CVPR)}, 2017.

\bibitem{Isola2017}
Phillip Isola, Jun-Yan Zhu, Tinghui Zhou, and Alexei~A. Efros.
\newblock Image-to-image translation with conditional adversarial networks.
\newblock In {\em IEEE Conference on Computer Vision and Pattern Recognition
  (CVPR)}, 2017.

\bibitem{Jin2017}
Zhiwei Jin, Juan Cao, Yongdong Zhang, Jianshe Zhou, and Qi Tian.
\newblock {Novel visual and statistical image features for microblogs news
  verification}.
\newblock {\em IEEE Trans. on Multimedia}, 19(3):598--608, 2017.

\bibitem{Johnson2016}
Justin Johnson, Alexandre Alahi, and Li Fei-Fei.
\newblock Perceptual losses for real-time style transfer and super-resolution.
\newblock In {\em European conference on computer vision}, pages 694--711,
  2016.

\bibitem{Karras2018}
Tero Karras, Timo Aila, Samuli Laine, and Jaakko Lehtinen.
\newblock {Progressive Growing of GANs for Improved Quality, Stability, and
  Variation}.
\newblock In {\em International Conference on Learning Representations}, 2018.

\bibitem{Karras2019}
Tero Karras, Samuli Laine, and Timo Aila.
\newblock A style-based generator architecture for generative adversarial
  networks.
\newblock In {\em IEEE Conference on Computer Vision and Pattern Recognition
  (CVPR)}, pages 4401--4410, 2019.

\bibitem{karras2020analyzing}
Tero Karras, Samuli Laine, Miika Aittala, Janne Hellsten, Jaakko Lehtinen, and
  Timo Aila.
\newblock {Analyzing and improving the image quality of StyleGAN}.
\newblock In {\em CVPR}, pages 8110--8119, 2020.

\bibitem{Kirchner2009}
Matthias Kirchner and Rainer B{\"o}hme.
\newblock {Synthesis of color filter array pattern in digital images}.
\newblock In {\em Media Forensics and Security}, 2009.

\bibitem{Kirchner2015}
Matthias Kirchner and Thomas Gloe.
\newblock Forensic camera model identification.
\newblock In T.S. Ho and S. Li, editors, {\em Handbook of Digital Forensics of
  Multimedia Data and Devices}, pages 329--374. Wiley-IEEE Press, 2015.

\bibitem{Liu2017}
Yanpei Liu, Xinyun Chen, Shanghai~Jiao Tong, Chang Liu, and Dawn Song.
\newblock Delving into tranferable adversarial examples and black-box attacks.
\newblock In {\em International Conference on Learning Representations}, 2017.

\bibitem{Lukas2006}
Jan Luk{\`{a}}{\v{s}}, Jessica Fridrich, and Miroslav Goljan.
\newblock Digital camera identification from sensor pattern noise.
\newblock {\em IEEE Trans. Inf. Forensics Security}, 1(2):205--214, 2006.

\bibitem{Lyu2014}
Siwei Lyu, Xunyu Pan, and Xing Zhang.
\newblock {Exposing Region Splicing Forgeries with Blind Local Noise
  Estimation}.
\newblock {\em International Journal of Computer Vision}, 110(2):202--221,
  2014.

\bibitem{Madry2017}
Aleksander Madry, Aleksandar Makelov, Ludwig Schmidt, Dimitris Tsipras, and
  Adrian Vladu.
\newblock Towards deep learning models resistant to adversarial attacks.
\newblock {\em arXiv preprint arXiv:1706.06083}, 2017.

\bibitem{Marra2018gan}
Francesco Marra, Diego Gragnaniello, Giovanni Poggi, and Luisa Verdoliva.
\newblock {Detection of GAN-Generated Fake Images over Social Networks}.
\newblock In {\em IEEE Conference on Multimedia Information Processing and
  Retrieval (MIPR)}, pages 384--389, 2018.

\bibitem{Marra2018vulnerability}
Francesco Marra, Diego Gragnaniello, and Luisa Verdoliva.
\newblock {On the vulnerability of deep learning to adversarial attacks for
  camera model identification}.
\newblock {\em Signal Processing: Image Communication}, 65, 2018.

\bibitem{Marra2019}
Francesco Marra, Diego Gragnaniello, Luisa Verdoliva, and Giovanni Poggi.
\newblock {Do GANs leave artificial fingerprints?}
\newblock In {\em IEEE Conference on Multimedia Information Processing and
  Retrieval (MIPR)}, pages 506--511, 2019.

\bibitem{Matern2019}
Falko Matern, Christian Riess, and Mark Stamminger.
\newblock Exploiting visual artifacts to expose deepfakes and face
  manipulations.
\newblock In {\em IEEE WACV Workshop on Image and Video Forensics}, 2019.

\bibitem{miyato2018spectral}
Takeru Miyato, Toshiki Kataoka, Masanori Koyama, and Yuichi Yoshida.
\newblock Spectral normalization for generative adversarial networks.
\newblock {\em arXiv preprint arXiv:1802.05957}, 2018.

\bibitem{Natsume2018}
Ryota Natsume, Tatsuya Yatagawa, and Shigeo Morishim.
\newblock {RSGAN: Face Swapping and Editing using Face and Hair Representation
  in Latent Spaces}.
\newblock In {\em ACM SIGGRAPH}, 2018.

\bibitem{Neves2020GANprintR}
João~C. Neves, Ruben Tolosana, Ruben Vera-Rodriguez, Vasco Lopes, Hugo
  Proença, and Julian Fierrez.
\newblock Ganprintr: Improved fakes and evaluation of the state of the art in
  face manipulation detection.
\newblock {\em IEEE Journal of Selected Topics in Signal Processing},
  14(5):1038--1048, 2020.

\bibitem{Nirkin2019}
Yuval Nirkin, Yosi Keller, and Tal Hassner.
\newblock {FSGAN: Subject Agnostic Face Swapping and Reenactment}.
\newblock In {\em IEEE International Conference on Computer Vision}, Oct. 2019.

\bibitem{Poursaeed2018}
Omid Poursaeed, Isay Katsman, Bicheng Gao, and Serge Belongie.
\newblock {Generative Adversarial Perturbations}.
\newblock In {\em IEEE Conference on Computer Vision and Pattern Recognition
  (CVPR)}, pages 4422--4431, 2018.

\bibitem{Qian2019}
Shengju Qian, Kwan-Yee Lin, Wayne Wu, Yangxiaokang Liu, Quan Wang, Fumin Shen,
  Chen Qian, and Ran He.
\newblock {Make a Face: Towards Arbitrary High Fidelity Face Manipulation}.
\newblock In {\em IEEE International Conference on Computer Vision}, 2019.

\bibitem{Roessler2019}
Andreas R{\"{o}}ssler, Davide Cozzolino, Luisa Verdoliva, Christian Riess,
  Justus Thies, and Matthias Nie{\ss}ner.
\newblock {FaceForensics++: Learning to Detect Manipulated Facial Images}.
\newblock In {\em International Conference on Computer Vision (ICCV)}, 2019.

\bibitem{Schroff2015}
Florian Schroff, Dmitry Kalenichenko, and James Philbin.
\newblock Facenet: A unified embedding for face recognition and clustering.
\newblock In {\em IEEE Conference on Computer Vision and Pattern Recognition
  (CVPR)}, 2015.

\bibitem{Song2018}
Yang Song, Rui Shu, Nate Kushman, and Stefano Ermon.
\newblock {Constructing Unrestricted Adversarial Examples with Generative
  Models}.
\newblock In {\em Conference on Neural Information Processing Systems (NIPS)},
  2018.

\bibitem{Szegedy2016}
Christian Szegedy, Vincent Vanhoucke, Sergey Ioffe, Jon Shlens, and Zbigniew
  Wojna.
\newblock Rethinking the inception architecture for computer vision.
\newblock In {\em IEEE Conference on Computer Vision and Pattern Recognition
  (CVPR)}, 2016.

\bibitem{Thies2016}
Justus Thies, Michael Zollh{\"o}fer, Marc Stamminger, Christian Theobalt, and
  Matthias Nie{\ss}ner.
\newblock {Face2Face: Real-Time Face Capture and Reenactment of RGB Videos}.
\newblock In {\em IEEE Conference on Computer Vision and Pattern Recognition},
  pages 2387--2395, June 2016.

\bibitem{Tuama2016}
Amel Tuama, Fr{\'e}d{\'e}ric Comby, and Marc Chaumont.
\newblock Camera model identification with the use of deep convolutional neural
  networks.
\newblock In {\em IEEE Workshop on Information Forensics and Security (WIFS)},
  pages 1--6, 2016.

\bibitem{Verdoliva2020review}
Luisa Verdoliva.
\newblock Media forensics and deepfakes: An overview.
\newblock {\em IEEE Journal of Selected Topics in Signal Processing},
  14(5):910--932, 2020.

\bibitem{Wang2019dec}
Sheng-Yu Wang, Oliver Wang, Richard Zhang, Andrew Owens, and Alexei~A Efros.
\newblock {CNN-generated images are surprisingly easy to spot... for now}.
\newblock In {\em CVPR}, 2020.

\bibitem{Wang2018}
Ting-Chun Wang, Ming-Yu Liu, Jun-Yan Zhu, Andrew Tao, Jan Kautz, and Bryan
  Catanzaro.
\newblock {High-resolution image synthesis and semantic manipulation with
  conditional GANs}.
\newblock In {\em IEEE Conference on Computer Vision and Pattern Recognition},
  2018.

\bibitem{Wang2019}
Xiaosen Wang, Kun He, and John~E. Hopcroft.
\newblock {AT-GAN: A Generative Attack Model for Adversarial Transferring on
  Generative Adversarial Nets}.
\newblock {\em arXiv preprint arXiv:1904.07793v3}, 2019.

\bibitem{Wu2019relgan}
Po-Wei Wu, Yu-Jing Lin, Che-Han Chang, Edward~Y. Chang, and Shih-Wei Liao.
\newblock Relgan: Multi-domain image-to-image translation via relative
  attributes.
\newblock In {\em IEEE International Conference on Computer Vision (ICCV)},
  2019.

\bibitem{Xiao2018}
Chaowei Xiao, Bo Li, Jun-Yan Zhu, Warren He, Mingyan Liu, and Dawn Song.
\newblock {Generating Adversarial Examples with Adversarial Networks}.
\newblock In {\em International Joint Conference on Artificial Intelligence},
  2018.

\bibitem{Yang2019}
Xin Yang, Yuezun Li, Honggang Qi, and Siwei Lyu.
\newblock {Exposing GAN-synthesized Faces using Landmark Locations}.
\newblock In {\em ACM Workshop on Information Hiding and Multimedia Security},
  pages 113--118, 2019.

\bibitem{Zhang2019}
Xu Zhang, Svebor Karaman, and Shih-Fu Chang.
\newblock {Detecting and Simulating Artifacts in GAN Fake Images}.
\newblock In {\em IEEE Workshop on Information Forensics and Security (WIFS)},
  2019.

\bibitem{Zhu2017}
Jun-Yan Zhu, Taesung Park, Phillip Isola, and Alexei~A. Efros.
\newblock Unpaired image-to-image translation using cycle-consistent
  adversarial networks.
\newblock In {\em IEEE International Conference on Computer Vision (ICCV)},
  2017.

\end{thebibliography}
}


\end{document}